\documentclass[10pt,twocolumn]{IEEEtran}
\usepackage{graphicx}           
\usepackage{amsmath,amsfonts}
\usepackage{times,epsfig}
\usepackage{psfrag}
\usepackage{subfigure}
\usepackage{stfloats}
\usepackage{amssymb}
\usepackage{multirow}
\usepackage[justification=centering]{caption}
\usepackage{cite}
\usepackage{CJK}
\usepackage{url}
\usepackage{caption}
\usepackage{mathrsfs}
\usepackage{bm}
\usepackage{enumerate}
\usepackage{color,xcolor}
\usepackage{colortbl}
\usepackage{multirow}
\usepackage{array}
\usepackage{algorithmic}
\usepackage[lined,ruled,linesnumbered]{algorithm2e}

\begin{document}
\pagenumbering{arabic}
\bibliographystyle{ieeetr}

\title{Service Reservation and Pricing for Green Metaverses:  A Stackelberg Game Approach}
 
\author{ Xumin Huang,  Yuan Wu, \emph{Senior Member, IEEE}, Jiawen Kang, Jiangtian Nie,\\ Weifeng Zhong,  Dong In Kim, \emph{Fellow, IEEE}, and Shengli Xie, \emph{Fellow, IEEE}
\IEEEcompsocitemizethanks{
	\IEEEcompsocthanksitem Xumin Huang is with School of Automation, Guangdong University of Technology, Guangzhou 510006, China, and also with State Key Laboratory of Internet of Things for Smart City, University of Macau, Macau, China (e-mail: huangxu\_min@163.com).
     Jiawen Kang, Weifeng Zhong and Shengli Xie are with School of Automation, Guangdong University of Technology, Guangzhou 510006, China.
      Yuan Wu is with State Key Laboratory of Internet of Things for Smart City, University of Macau,  Macau, China, and also with Department of Computer and Information Science, University of Macau,  Macau, China.  Jiangtian Nie is with School of Computer Science and Engineering, Nanyang Technological University, Singapore. 
       Dong In Kim is with the Department of Electrical and Computer Engineering, Sungkyunkwan University, Suwon 16419, South Korea.

}
}

\maketitle
\thispagestyle{empty}

\begin{abstract}
Metaverse enables users to communicate, collaborate and socialize with each other through their digital avatars. Due to the spatio-temporal characteristics, co-located users are served well by performing their software components in a collaborative manner such that a Metaverse service provider (MSP) eliminates redundant data transmission and processing, ultimately reducing the total energy consumption. The energy-efficient service provision is crucial for enabling the green and sustainable Metaverse. In this article,  we take an augmented reality (AR) application as an example to achieve this goal. Moreover, we study an economic issue on how the users reserve offloading services from the MSP and how the MSP determines an optimal charging price since each user is rational to decide whether to accept the offloading service by taking into account the monetary cost. A single-leader multi-follower Stackelberg game is formulated between the MSP and users while each user optimizes an offloading probability to minimize the weighted sum of time, energy consumption and monetary cost.  Numerical results show that our scheme achieves energy savings and satisfies individual rationality simultaneously compared with the conventional schemes. Finally, we identify and discuss open directions on how several emerging technologies are combined with the sustainable green Metaverse.
\end{abstract}


\section{Introduction}
As a hypothesized next-generation Internet, Metaverse provides  a self-sustaining, hyper spatiotemporal, and three-dimensional (3D)  immersive virtual shared space for users in many areas, e.g., business, entertainment, training and education. The Metaverse enables users to communicate, collaborate, and socialize with each other by using digital avatars as virtual identities \cite{9944868}. To enhance user experience in Metaverse applications,  innovative information and communication technologies are exploited. As one of the cornerstone technologies of Metaverse, XR including virtual reality (VR), augmented reality (AR) and mixed reality (MR) is applied to provide extremely interactive and fully immersive experience for Metaverse users.  5G/B5G enhances network-wide data communications by providing higher throughput, very low latency and ubiquitous connectivity \cite{yu20226g}. Digital twin creates a digital representation of physical objects and systems to generate a digital mirror of the real world, and aims to achieve real-time physical-virtual synchronization. Thanks to the above advancements,  Metaverse can host different virtual worlds as sub-Metaverses while supporting interconnection among the sub-Metaverses to maintain a unified Metaverse and guarantee highly interactive immersive experience for the users  anywhere and anytime. Nowadays, a variety of Metaverse services are provided by different Metaverse service providers (MSPs). For example, users can have a VR chat via Oculus Rift,  play interesting VR/AR games such as  Horizon worlds and Pokémon Go, and enjoy social activities in Minecraft and AR experience in autonomous driving. At present, the most popular way of the users to experience the Metaverse is via diverse AR/VR applications. According to current market research and statistics,  AR/VR games are the most popular Metaverse application in the global market, which is predicted to reach  8 trillion dollars by 2030. Thus, we are motivated to investigate the AR application as a representative application of Metaverse. However, the number of global connected Internet of Things (IoT) devices is predicted to reach 83 billion by 2024. With the increasing user density at the same location, the MSP could serve many co-located users at the same time. To avoid extra energy consumption, a few works have tried to seek energy-efficient operations, e.g., caching proper data \cite{9252973}  and controls the frequency of data transmission and processing.

In this article, we study optimization methods on the MSP side to serve the co-located users in an energy-efficient way and achieve the green Metaverse. Users can access a sub-Metaverse with lower latency by requesting offloading services from the MSP that has rented available bandwidth and computing resources at the network edge.  To improve processing efficiency, co-located users can be elaborately served together. Due to the same or similar surrounding environment,  the users may submit same input data, accept identical data processing, and finally share same output data during the offloading procedure. For example, the nearby vehicles have identical 3D map information in a specific region. Thus, we explore effective operations between the MSP and users to eliminate redundant data transmission and processing.  Collaborative efforts between both parties facilitate the Metaverse service provision with lower energy consumption and this is helpful for the progress of the green Metaverse.

In this regard, we take an AR application as an example. Previous works \cite{7906521,8690922} have shown that software components of the users in the same surrounding environment  can be performed in a collaborative manner for energy savings.  The basic components of a user in the AR application include video source (fetching original video frames), tracker (estimating the position of the user during the movements),  mapper (building a map of the environment based on the  extracted feature points), object recognizer (identifying the known objects), and render (rendering virtual objects on the screen). The co-located users can offload their common content and service data (e.g., software components) to an identical MSP. When detecting the same surrounding environment, the MSP further utilizes the collaborative nature of their trackers, mappers and object recognizers to collect the commonly utilized input data, perform the identical workloads once, and share the basic output data with all users. This reduces communication overheads and computing workloads generated in the sub-Metaverse. Both the time and energy consumptions are saved for the Metaverse service, especially when dealing with massive co-current requests from the co-located users, which is important for enabling the green Metaverse.

In this article, we study the green AR service provision and further investigate an economic issue related to how the MSP serves the co-located users in an energy-efficient manner while controlling their offloading willingness by a pricing scheme. The MSP aims to maximize the expected revenue  under a probabilistic and fair offloading scenario, while each user makes a probabilistic offloading decision to minimize the weighted sum of time, energy consumption and monetary cost. Accounting for the fairness issue, the MSP adopts a uniform charging price and uniform resource allocation among the offloading users. To model the scenario, we formulate a Stackelberg game between the MSP and users when the MSP collects sufficient user profiles and predicts their responses to different charging prices. Then the MSP plays as the game leader who has the privilege over the users and takes actions first, and users play as the game followers who determine their optimal responses according to the MSP's pricing strategy. We solve the Stackelberg game to determine the service reservation and pricing on the user and MSP sides, respectively. Finally, we discuss open directions on how emerging technologies such as 5G/B5G, digital twin, wireless energy transfer, blockchain and artificial intelligence (AI) can be leveraged for enabling the green Metaverse.

The main contributions of this article are summarized as follows:

\begin{itemize}
	\item We exploit collaborative awareness of the co-located users to fulfill the green Metaverse. We take the AR application as an example to illustrative the simultaneous processing of software components of the users.

	\item We employ a Stackelberg game to tackle the service reservation and pricing problem.  Probabilistic behavior, fairness and individual rationality are considered in our game model.
	
	\item We outline research challenges of the green Metaverse and potential solutions to key issues in networking, modeling, energy supply, decentralized economy ecosystem and AI applications.
	
	\end{itemize}

\begin{figure*}
	\centering
	\includegraphics[width=1.0\textwidth]{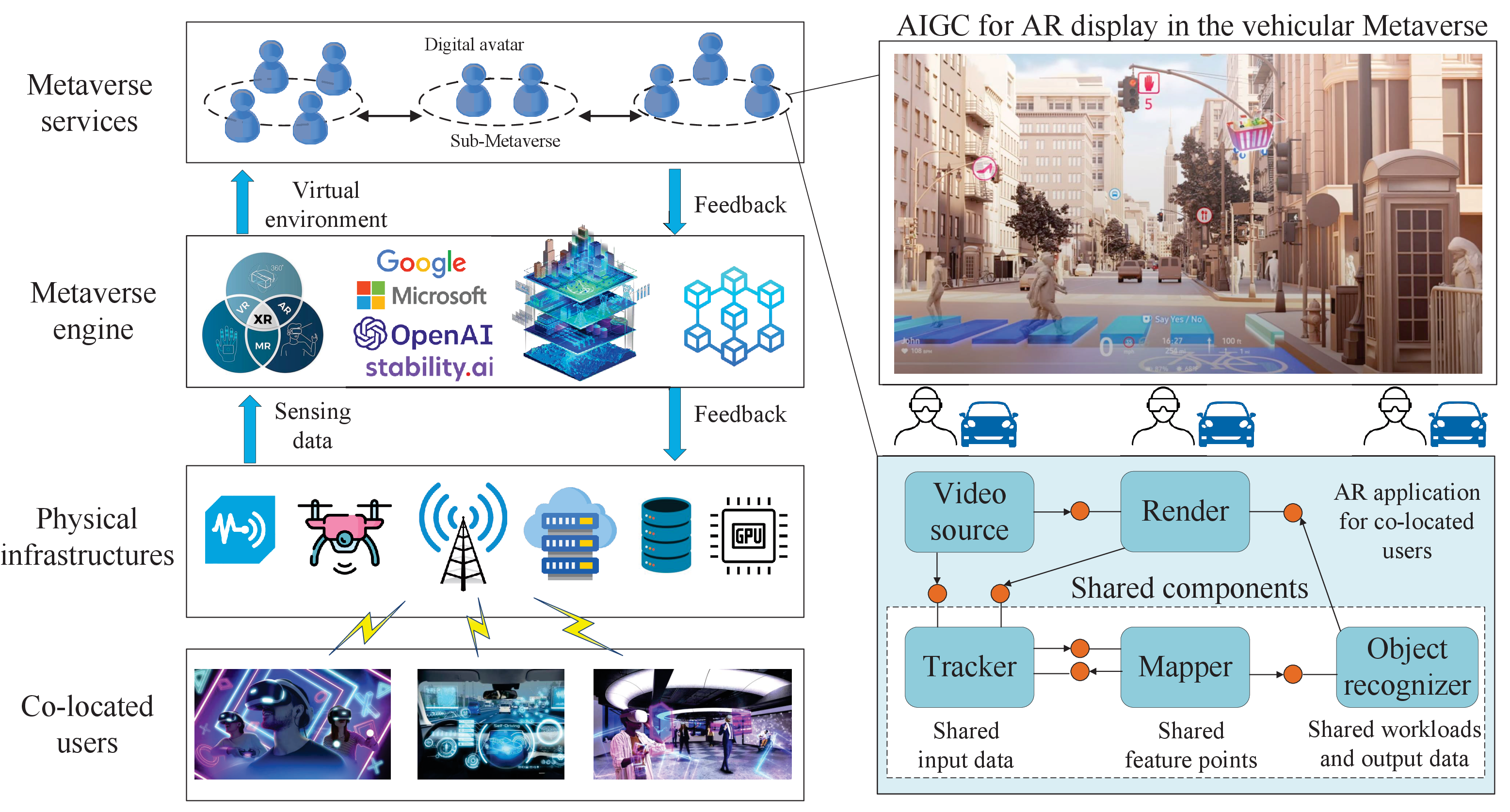}
	\caption{The Framework of the green Metaverse}
	\label{system}
\end{figure*}

\section{Green Metaverses and Related Research}
\subsection{Related Research on Metaverses}

Research efforts have been devoted to different optimization schemes for the Metaverse. Metaverse users necessitate available bandwidth and CPU frequencies to run Metaverse applications and the Metaverse consumes network resources to maintain different virtual objects for the users. This leads to different resource allocation problems among the users. In \cite{9838736, 9973495}, the bandwidth allocation problem was studied by using auction or game theories when different users access the Metaverse over different wireless channels. The computing resource allocation problem was investigated when a Metaverse service requires computing-intensive tasks such as real-time rending tasks. In \cite{9880566}, the concept of vehicular Metaverse was introduced and code distributed computing was adopted to employ nearby vehicles with underutilized computing resources to execute rendering tasks for a vehicular user. A Stackelberg game based incentive mechanism was further presented for the computing resource trading. To allocate more network resources to virtual objects in which users were more interested, an attention-aware algorithm was developed in \cite{9999298}. Besides, up-to-date status data of physical objects were rather crucial for efficient synchronization of digital twin models of different MSPs \cite{9865226}. The synchronization intensity control problem among multiple MSPs was studied according to a Stackelberg differential game.

Furthermore, some researches have begun to implement prototypes and present case studies of the Metaverse. The combination of autonomous vehicles and Metaverse was conceived to help upgrade the current personal mobility pattern \cite{deveci2022}. The potential of autonomous vehicles in the Metaverse was discussed to enrich the personal mobility and extend the transportation system. A blockchain-driven university Metaverse was developed to make the social experiments on the on-campus students more convenient \cite{duan2021}. The key enabling technologies for edge-driven Metaverse were summarized and a case study regarding virtual smart city was provided \cite{9815180}. 

However, previous works neglected co-located users in the Metaverse applications, which have the same location in the real world and simultaneously access the Metaverse.  For example, a user may invite the family or friends to play a multi-participant AR/VR game in the living room and the co-located users engage in a number of immersive experiences and activities together in the Metaverse. Currently, there is limited research on optimizing services for the co-located users
and how to serve them efficiently while conserving energy remains an unaddressed issue. Due to the same location, the co-located users can be elaborately served by processing the same or similar data. This motivates us to propose an energy-saving approach to enable an MSP to serve the co-located users with less data transmission and workload processing, which is helpful to reduce the total energy consumption and beneficial for the achievement of the green Metaverse. Besides, the MSP should consider how to derive an optimal charging price for the users while considering the individual rationality.
\subsection{System Model for Green Metaverses}
A hierarchical framework of the Metaverse is presented in Fig.~\ref{system}, which includes the human world, physical world and digital world.  In the human world,  co-located users equipped with smart wearable devices can simultaneously access the Metaverse. The physical world includes massive deployment of physical infrastructures for data perception, transmission and processing. The digital world refers to the implementation of the Metaverse, where diverse enabling technologies are jointly applied as the Metaverse engine and interconnected sub-Metaverses accommodate a variety of Metaverse services. 
\begin{itemize}
	\item \emph{Human world}: The Metaverse permits each user in the human world to create a customized digital avatar that is freely handled to play, work, communicate and interact with surrounding virtual objects or other digital avatars. The digital avatars are controlled through smart wearable devices such as head-mounted display, VR/AR goggle, haptic gloves and brain-computer interface. 
	\item \emph{Physical world}: For fully coherent and immersive experience,  high-quality interactions between the users and digital avatars depend on multi-source environmental data acquisition, ubiquitous connection, real-time data processing and visualization. This necessitates massive deployments of fixed and mobile sensors, e.g.,  unmanned aerial vehicle, communication, computing and storage infrastructures in the physical world.  With the large-scale applications of edge/cloud computing, offloading computation-intensive application tasks to nearby edge/cloud servers is adopted to break through the hardware limitations of the user devices.
	\item \emph{Digital world}: A virtual world is created to mirror the real world based on the fine-grained data. The virtual world is maintained and updated on demand with the help of the Metaverse engine, which refers to a variety of enabling technologies such as XR, AI, 3D simulation and blockchain. The XR technology is convinced to be the core of the Metaverse and blockchain builds a decentralized economy ecosystem for the Metaverse. AI provides techniques and applications for the foundation and development of the Metaverse, but also  enhances the decision-making capability to intelligently run the Metaverse applications. Particularly, AI-generated content (AIGC) can provide a promising generative tool to produce personalized content based on user-oriented requirements and make the virtual worlds more vivid for the Metaverse. Many AI leading companies like OpenAI, Stability AI, Google and Microsoft make important contributions for the technology development and wide applications of AIGC. For example, we introduce a potential integration of AIGC with the AR-empowered vehicular Metaverse, where AIGC is applied to produce and render virtual objects in the AR display according to audio inputs of a driver. According to differnt application scenarios, different sub-Metaverses are created to provide virtual environments, objects and services for the users. The real-time interconnection among different sub-Metaverses ensures the seamless user experience in the unified Metaverse. 
\end{itemize}

\begin{figure}[t!]
	\centering
	\includegraphics[width=0.45\textwidth]{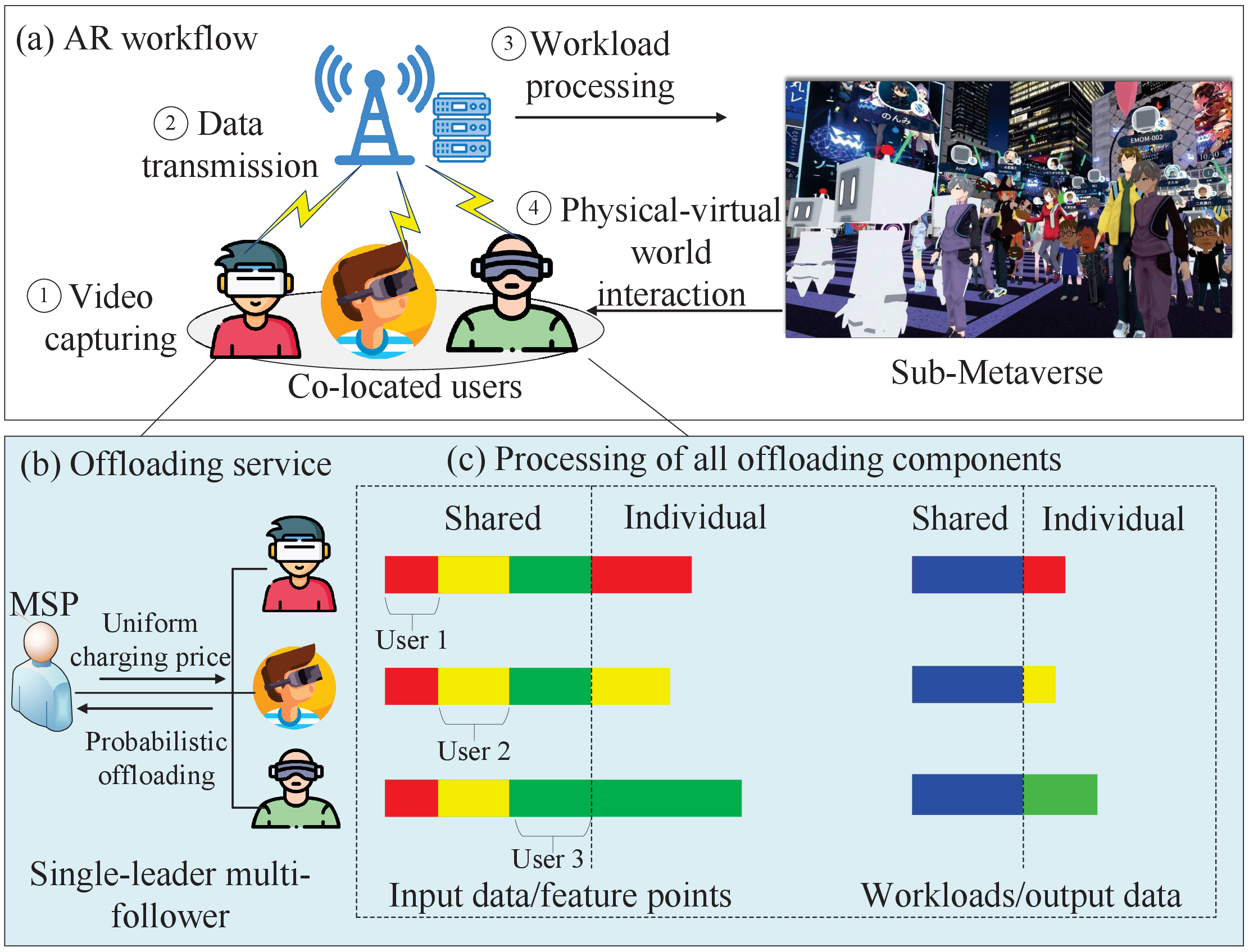}
	\caption{The energy-efficient AR application among the co-located users}
	\label{component}
\end{figure}

It is noticed that the potential collaborative nature of a Metaverse service, e.g., AR, can be explored from special users, e.g.,  co-located users. In Fig.~2(a), we introduce the basic workflow of the AR application. The users first capture their video frames by using cameras, and can choose to offload their software components with necessitated data to an edge server of the MSP. Then the MSP performs the software components to render and encode the received video frames, which are transmitted to the corresponding users. With the help of the smart wearable devices, the users decode and display the video frames to access a sub-Metaverse and enjoy physical-virtual world interactions. 

When serving multiple co-located AR users, an MSP can process the same or similar data for them. The previous works have proposed the shared components among the co-located AR users consisting of their trackers, mappers and object recognizers. The software components can be offloaded to an edge server of the MSP for prompt processing, as shown in Fig.~2(b). To facilitate the simultaneous processing of the offloading components, the MSP collects a part of the input data from each tracker to generate the commonly utilized input data among all users. Besides, the MSP recognizes some same objects for the users in the same or similar surrounding environment such that the workloads of computing the same feature points and objects can be performed once in the mapper and object recognizer respectively, and associated outcome is conveniently shared with the users. As shown in Fig.~2(c), we exploit the shared components and describe the energy-efficient AR application among the co-located users as follows.  
\begin{itemize}
	\item \emph{Shared input data}: Each user submits a part of captured video frames to generate the shared input data commonly utilized among the users, which is transmitted to their object recognizers.
	\item \emph{Shared feature points}: When reconstructing feature points of the video frames submitted by the users, computation results of some feature points in the same surrounding environment are cached in advance and shared to the associated users. This allows each mapper to be performed with partial workloads.
	\item \emph{Shared workloads and output data}:  For the co-located users, the identical objects in the same area are selected to be recognized once. The output data regarding the same objects is shared to the corresponding users. 
\end{itemize}
When serving the offloading users that determine to offload their software components,  the MSP coordinates the execution of their  trackers, mappers and object recognizers to eliminate the redundant data transmission and processing, and finally saves both the time and energy consumptions of the Metaverse service. The network-wide energy consumption is reduced, which is significant for achieving the green Metaverse.

Low network latency is critical to maintaining an immersive user experience. The MSP stimulates a user to offload the software components to the edge server for prompt execution. For the user, the latency of the offloading service is majorly influenced by the number of computing resources allocated to process the offloading task in the edge server. To incentivize more users to accept the offloading service, the MSP promises a minimal number of computing resources per user and specifies a uniform charging price for all users.  The minimal number is calculated by dividing the total number available on the edge server by all users. The information is recorded to a service-level agreement that is delivered to all users. According to the service-level-agreement, the users looking forward to an improved immersive user experience in the Metaverse may choose to accept the offloading service.

\section{Stackelberg Game based Service Reservation and Pricing}
\subsection{Stackelberg Game Model}
In this article, we study  how the co-located AR users reserve their offloading services from an MSP that has rented a dedicated edge server with available computing resources to provide computation offloading, and accordingly, how the MSP  optimizes the charging price for the users. A user in the sub-Metaverse is rational to decide whether to offload the software components to the MSP, according to the charging price of the offloading service. 

To maximize the economic benefits of the service provision, we propose a Stackelberg game  between the MSP and users and enable the MSP's domination over the entire game duration. A Stackelberg game refers to a two-stage game between two types of game players, where one type of the players named by game leaders take action first, and the other type of the players named by the game followers make decisions after the game leaders. Each user determines whether to accept an offloading service from the MSP, according to the pricing strategy. According to historical interactions with personalized digital avatars of the users, the MSP can collect sufficient user profiles and obtain prior knowledge of the users. The MSP predicts their responses to different charging prices in advance. Thus, the MSP who has the informational advantage is suitable to play as a game leader that has the privilege over the users and takes action first. Then users play as game followers who determine their optimal responses according to the MSP's pricing strategy. Based on the above consideration, we employ a single-leader multi-follower Stackelberg game in Fig.~2(b) to formulate the strategic interactions between the MSP and users. Moreover, we consider the following feasible assumptions for the offloading services.
\begin{itemize}
	\item \emph{Probabilistic offloading}: When experiencing a virtual world, users face uncertain demands such as resource demand \cite{ng2022unified}. Due to the uncertainty, users tends to make probabilistic offloading decisions instead of deterministic offloading decisions. As a result, each user decides to offload the software components in a probabilistic manner. 
	\item \emph{Fairness consideration}: Since the offloading users are located in the same area, they can easily compare charging prices and computing resource allocation results with each other. For the fairness, we let the MSP adopt the uniform pricing strategy for them and uniform computing resource allocation among them. 
	\item \emph{Leader-follower relationship}: The MSP has the prior knowledge of the users and recognizes their strategies.  This enables the MSP to be in a more advantageous position than the users. Thus, a leader-follower relationship is established between the MSP and users.
\end{itemize}

in the Stackelberg game, we define objective functions of a user and the MSP.  In the sub-Metaverse, a user has a probability $\alpha$ of offloading the software components to the MSP. Then the probability of locally executing the software components is $1-\alpha$. The user optimizes $\alpha$ to reduce the expected total cost, which is  relevant to the  time consumption cost, energy consumption cost and monetary cost.  In the local execution case,  completing workloads of the user are equal to  the sum of the workloads of the tracker, mapper, object recognizer and render. When knowing the local computing frequency and some hardware parameters  such as the effective capacitance coefficient, we can calculate the task completion time and energy consumption cost. 

In the following, we describe the offloading case. An example of three users with shared components in computation offloading is shown in Fig.~\ref{component}(c). We provide the quantitative analysis on the shared components as follows.
\begin{itemize}
	\item \emph{Tracker}: When offloading a tracker to the MSP, each offloading user is first required by the MSP to submit a part of the input data to generate the commonly utilized input data among all offloading users, which refers to the shared input data. Then the residual  input data  of the  offloading user becomes individual input data. 
	For each offloading user, the input data transmission time consists of two parts: waiting time of collecting the shared input data and transmission time of the individual input data. 
	\item \emph{Mapper}: When offloading a mapper,  the MSP only needs to complete a part of the original workloads for an offloading user. The workloads of the offloading  mapper are related to the proportion of its feature points that are pre-processed and the computing results are cached. The calculation method of this proportion is presented in  \cite{8690922}. 
	\item \emph{Object recognizer}: When offloading an object recognizer, workloads of recognizing  a part of the same objects are specified by the MSP.  
	The MSP completes the workloads of recognizing the same objects among all offloading users by the entire computing frequency $F$. By adding the workloads of a render, we calculate the individual workloads of each offloading user. Note that the individual workloads are performed by the uniform computing frequency $F/\sum{{\alpha}}$. For each offloading user, the workload processing time is consumed to first process the shared workloads once, and then process the individual workloads. Finally, the output data of recognizing the same objects, i.e., shared output data, is broadcast to all offloading users while the residual output data, i.e., individual output data, is transmitted to associated offloading users.  For each offloading user, the output data transmission time  consists of two parts: broadcast time of transmitting the shared output data and transmission time of the individual output data.
\end{itemize}
Given the transmit and receive power, we can calculate the energy consumption cost of each offloading user. Until now, we calculate the task completion time and energy consumption cost of each offloading user in the offloading case.

In this article, we introduce an expected total cost function of the user, which is expressed as a weighted sum of the time consumption cost, energy consumption cost and monetary cost. The time consumption cost is measured by introducing the penalty cost per unit time. The penalty parameter is an application-centric parameter, which is related with the user-defined preference for tolerating the latency of the immersive experience. The monetary cost refers to the uniform charging price for all users denoted by $p$. Each user requires necessary auxiliary information from the MSP to assist in the decision-making procedure. The MSP aims to maximize the expected revenue $p\sum {{\alpha}}$. In the proposed Stackelberg game, the MSP plays as a game leader and has the ability to move first and determines the charging price, since it possesses the prior knowledge about the users' possible reactions to the actions taken by the game leader.

\subsection{Stackelberg Equilibrium Analysis}
To obtain the solution of the formulated Stackelberg game, we seek the Stackelberg equilibrium, at which the MSP can maximize the expected revenue given the best responses of the users that minimize their expected total cost. At this equilibrium, neither any game leader nor any game follower can improve the individual utility by unilaterally changing the strategy. We adopt the backward induction method to analyze the Stackelberg equilibrium. We derive the best response of each user by taking the first and second order derivatives of the cost function with respect to the offloading probability, and realize that the function is strictly convex with respect to the offloading probability.

We can derive the optimal offloading probability of any user according to the first-order optimality condition. First, we derive the aggregated offloading demand of all users by summing all  first-order optimality conditions among all users. We get  the simplified expression  $\sum {{\alpha _i}}=\phi  - \theta p$, where $\phi$ and $\theta >0$. With the increase of $p$, the aggregated offloading demand decreases. This is consistent with the intuition. Second, we resolve $\alpha$ by substituting $\sum {{\alpha}}$ into the corresponding equation on its first-order optimality condition. After knowing the best responses of each user, we rewrite the expected revenue of the MSP as a typical concave quadratic function such that the optimal charging price is easily derived.

\section{Numerical Results}
We conduct the simulation experiments to evaluate the overall performance of our scheme for the green Metaverse. Referring to \cite{7906521}, we set similar simulation parameters for eight users and an MSP. In the AR application, the input data size, total workload and output data size of each user are identical.  The local computing frequency, data rate, transmit power, effective switched  capacitance coefficient and penalty cost per unit time of each user follow a uniform distribution with $U[1,2]$ GHz, $U[5,10]$ Mbps, $U[26,30]$ dbm, $U[5,10]\times 10^{-27}$ ${\rm{Watt}} \cdot {{\rm{s}}^3}/{\rm{cycl}}{{\rm{e}}^3}$  and $U[300,600]$ cents/second, respectively. For simplicity, we consider that for the offloading users, percentages of the reduction of the input data size, computation workloads and output data size  vary from 30\% to 40\% after performing their offloading components in a collaborative manner. The computing frequency of the edge server belonging to the MSP is 10 GHz. The lower and upper limits of the charging price are 140 and 280 cents, respectively.

To demonstrate the superiority of our scheme, we compare our scheme with two baseline schemes. One of the baseline schemes is called by all local processing (ALP), which means that each task is locally processed. The other one is called by all task offloading (ATO), which means that each task is fully offloaded.  We compare the total energy consumption of completing all tasks and average cost of all users among the three schemes.  Compared with the ALP, our scheme greatly reduces the total energy consumption about 19\% and compared with the ATO, our scheme consumes slightly more energy since a part of the users finally choose to offload their software components. Clearly, the MSP can perform all software components of all users in a collaborative manner to achieve the maximal energy savings via the ATO. However, both ATO and ALP force each user to accept a uniform offloading decision and neglect the individual rationality of the users. The baseline schemes cannot enable the users to freely make the offloading decisions when they take into account the charging price of the MSP. In our scheme, each user can rationally optimize the probabilistic offloading decision to minimize an expected total cost function related with the weighted sum of the time consumption, energy consumption and monetary cost. In this regard, our scheme obtains a minimal average cost of all users compared with the ALP and ATO. The cost decline reaches more than 20\% and 4\%, respectively. As a result, the comparison shows that our scheme has an advantage in reducing the network-wide energy consumption for realizing the green Metaverse while guaranteeing the feasibility of the pricing strategy of the MSP by considering the individual rationality of the users.

\begin{figure}
	\centering
	\includegraphics[width=0.5\textwidth]{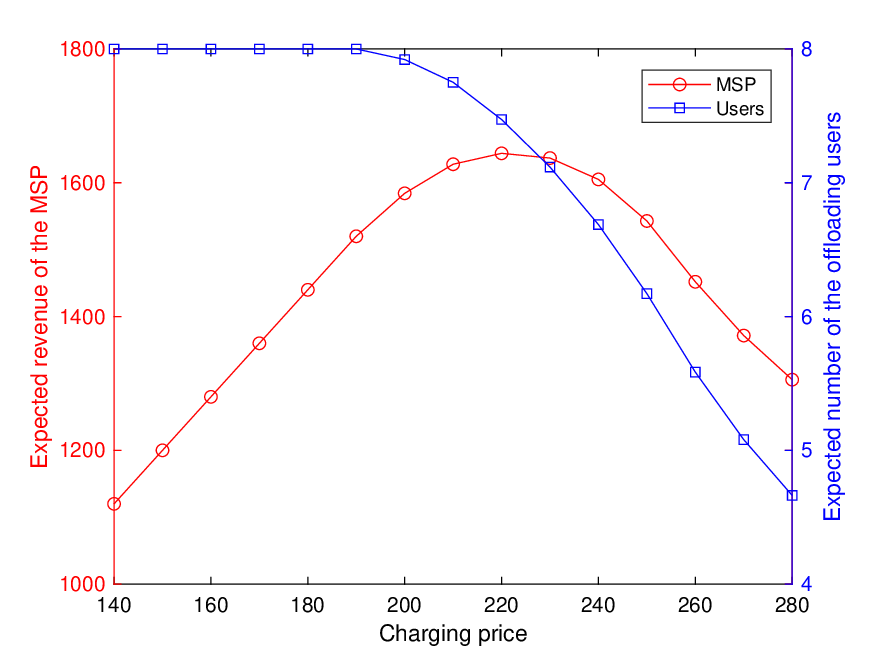}
	\caption{Strategies of the MSP and users with respect to different charging prices}
	\label{strategy}
\end{figure}

In addition, we compare the expected revenue of the MSP and the expected number of the offloading users with respect to different charging prices, as shown in Fig.~\ref{strategy}.  The expected revenue of the MSP varies with the change of the charging price. When the charging price is lower enough, each user would like to accept the offloading service. At this time, the expected revenue linearly increases with the increase in the charging price. When the charging price becomes high for some users, they are rational to choose a lower offloading probability to minimize the expected total cost. Then the expected number of the offloading users clearly decreases with the increase in the charging price. This is consistent with the intuition. From the figure, we also know that there exists a maximal expected revenue for the MSP when deriving the optimal charging price by tackling the  Stackelberg game.   

To summarize, the  numerical results show our scheme's superiority and efficiency in gaining the energy savings for achieving the green Metaverse while enabling the independent decision-making capabilities of both the users and MSP.

\begin{table}[t!]
	\renewcommand{\arraystretch}{2}
	\caption{Performance Comparison of Different Schemes}\label{source.properties} \centering \tabcolsep=5pt
	\begin{tabular}{|p{0.18\columnwidth}<{\centering}|p{0.3\columnwidth}<{\centering}|p{0.3\columnwidth}<{\centering}|}
		\hline
		~\textbf{Scheme} & \textbf{Total energy consumption } (J)& \textbf{Average cost (cents)} \\
		\hline
		ALP  & 273  & 8051 \\
		\hline
		ATO  & 174 & 6708 \\
		\hline
		Our scheme & 221 & 6434\\		
		\hline
	\end{tabular}
\end{table}

\begin{figure}
	\centering
	\includegraphics[width=0.5\textwidth]{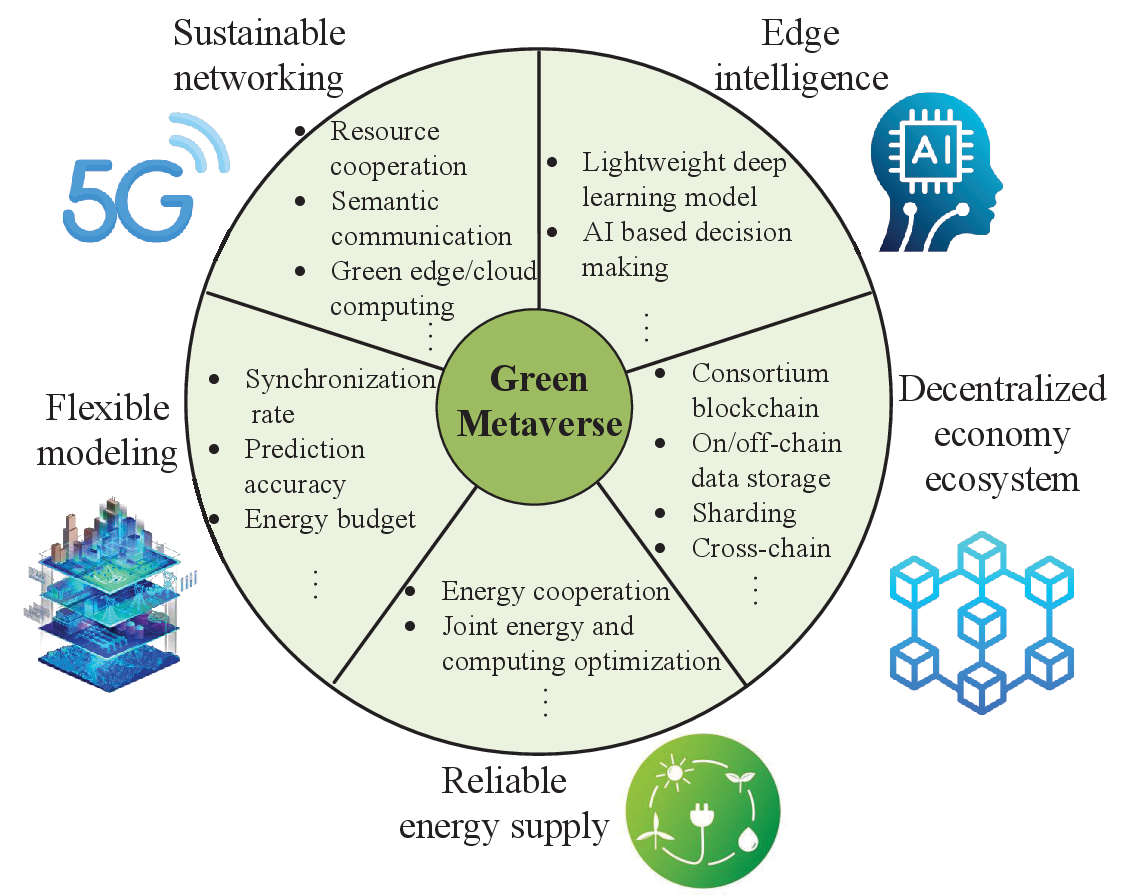}
	\caption{Future directions and focuses for the green Metaverse}
	\label{future}
\end{figure}

\begin{table*}[t]
	\renewcommand{\arraystretch}{2}
	\caption{Open challenges for the green Metaverse}\centering \tabcolsep=5pt
	\begin{tabular}{|p{0.25\columnwidth}<{\centering}|p{0.25\columnwidth}<{\centering}|p{0.65\columnwidth}<{\centering}|p{0.65\columnwidth}<{\centering}|}
		\hline
		\textbf{Key function} & \textbf{Core technology} & \textbf{Challenging issues} & \textbf{Potential solutions}\\
		\hline
		Sustainable networking &  5G/B5G & Interference among co-located users, co-current traffic, collaborative applications, resource provision  & Study interference control,  cooperation among base stations,  UAV deployment, semantic communication, cooperation among adjacent sub-Metaverses, and green edge/cloud  computing.       \\
		\hline
		Flexible modeling & Digital twin & Tradeoff between synchronization accuracy and synchronization rate&  Dynamically change the synchronization rate or size of each gathering data sample.\\
		\hline
		Reliable energy supply& Energy harvesting, wireless energy transfer& Intermittent and random nature of the energy harvesting process, UAV assisted wireless energy and data transfer & Propose proper energy cooperation, and study the joint optimization of UAV trajectory and computation offloading.\\
		\hline
		Decentralized economy ecosystem & Blockchain & Multiple blockchains, cross-chain, storage and processing overhead& Utilize lightweight consensus nodes and green consensus mechanisms, sharding technologies, and adaptive cross-chain communication schemes.\\
		\hline
		Edge intelligence  & AI & Computation and storage-intensive deep learning models, decision-making procedures & Reduce the sizes and complexity of current deep learning models, and make AI-enabled decisions.\\
		\hline
	\end{tabular}
\end{table*}

\section{Open Research Challenges}
With the dramatically increasing number of connected users in the Metaverse, it is extremely critical to seek more energy-efficient networking, modeling, energy supply and service delivery solutions for many immersive Metaverse services and applications. To enhance the experience quality for numerous users, multi-source data including visual, auditory, haptic and kinesthetic should be captured, transmitted, processed and shared at high throughput with wide-area access while eliminating  security and privacy risks. This requires sustainable networking, flexible modeling, reliable energy supply, decentralized economy ecosystem and edge intelligence. We propose the potential directions of future research towards green Metaverse in Fig~\ref{future}.  It should be note that the evolution of the Metaverse depends on the evolution of other technologies including 5G/B5G, digital twin, wireless energy transfer, blockchain and AI. There are some research challenges and potential solutions that can be considered for the green Metaverse, which are summarized in Table II. In the following, we identify the key issues and discuss potential solutions in detail.

\subsection{5G/B5G for Sustainable Networking}
The eMBB, nMTC and uRLLC technologies of 5G/B5G have great potential to meet the stringent communication requirements. However, the current solutions should be revised to achieve sustainable networking for the future Metaverse. The Metaverse will generate a vast number of co-current and co-located requests, leading to a sharp increase in energy consumption for communication and computation. For the green Metaverse, energy-efficient communication should tackle the interference among the co-located antennas according to distribution of the users. The joint optimization of the power control, channel selection and bandwidth allocation may be studied. To mitigate the co-current traffic offloading,  we propose the cooperation between non co-located  base stations, e.g, macro and micro base stations. Unmanned aerial vehicle (UAV) enabled  aerial  base stations can also be quickly deployed to extend the network coverage and capacity while maintaining infrastructure investment. Besides, semantic communication is a key technology of B5G that extracts the meaning from transmitted data as available information required for a respective task/goal, i.e., semantic information, and transmits the semantic information with a smaller data size instead of transmitting the entire original data.  For the Metaverse, semantic communication prevents raw data streams between the physical and virtual worlds from creating enormous data volume, and enables users to reduce data transmission load and energy consumption while improving the data transmission efficiency. The application of semantic communication is particularly attractive to  content dissemination services over bandwidth-limited wireless channels in the Metaverse.

From a computational viewpoint, more operations on the shared content delivery, data transmission and processing among  different MSPs are investigated to well serve the co-located users in different  Metaverse applications. For example, proper virtual objects are created, cached and maintained once and shared to the adjacent sub-Metaverses. More energy savings are obtained by reasonably performing some workloads once and re-using the same input/output data.  In addition, green Metaverse can be supported by harvesting enormous computing and storage resources available at the network edges to seamlessly provide Metaverse services for the users while enriching the resource supply. For example,  popular content can be cached in proper mobile devices that are willing to share their storage with others  \cite{9252973}. Neighbor vehicles with idle computing resources are scheduled to help implement a vehicular Metaverse \cite{9880566}. Then the efficient resource allocation strategies for the green edge/cloud computing are integrated to improve the resource utilization efficiency.

\subsection{Digital Twin for Flexible Modeling}
The success of the Metaverse is in line with the progress of digital twin. As a key enabling technology, digital twin is applied to create cloning of physical objects into the immersive 3D virtual world. 

For the targeted  physical objects, massive sensing of the statuses, multi-source data analysis and predictive modeling are periodically performed throughout their lifetime \cite{yu2022bi}. The routine operations clearly lead to a large amount of resources and energy. To restrict the energy consumption, the operation frequency is adjusted by changing the presetting synchronization rate, which has an important impact on the synchronization accuracy between physical and virtual objects. Note that the higher synchronization rate is beneficial to improve the synchronization accuracy but increase the energy consumption rate. To tackle the energy-performance tradeoff,  a fine-grained approach is to dynamically change the synchronization rate or size of each gathering data sample, according to the real-time energy state. In this way, the digital twin technology can be utilized well to produce an adaptive digital replica of the physical system to satisfy the specific predicting accuracy within an acceptable energy budget.
\subsection{Reliable Energy Supply}
To provide a reliable energy supply for the users, both the energy harvesting and wireless energy transfer techniques can be exploited. Nowadays, energy harvesting techniques are integrated into IoT and support IoT devices with long life-time and low maintenance. Then a variety of energy harvesting modules using renewable sources are embedded into Metaverse related user devices to offer an optional power supply for them. Due to the intermittent and random nature of the energy harvesting process, a potential solution to increasing the efficiency of energy utilization is to propose proper energy cooperation among different devices with different energy supplies. 
 
Besides, wireless energy transfer is applied as an effective tool to flexibly charge the above devices on demand. To this end, UAV assisted wireless energy transfer is proposed to let a UAV fly to a specific hovering location and provide wireless charging to the energy-constrained ground devices via the radio frequency energy transfer. At the same time, the UAV can provide computation offloading for the devices. To maximize the energy efficiency of the UAV, we can study a joint optimization problem of UAV trajectory and computation offloading subject to the energy consumption and task deadline constraints. 

\subsection{Blockchain for Decentralized Economy Ecosystem}
Blockchain has been adopted as the essential underlying technology to build a decentralized and sustainable economy ecosystem for future Metaverse where digital avatars can freely authenticate, value and trade their digital assets in the totally trustless environment. Different sub-Metaverses could establish different blockchains to manage the services. When users shuttle across different sub-Metaverses to experience a digital life, they need to access multiple blockchains and exchange their assets across different blockchains. To this end, cross-chain technology is necessitated to integrate multiple blockchains for the whole Metaverse. 

To attain sustainability for the Metaverse design in the blockchain environment, we discuss the potential solutions to the green Metaverse by running the blockchain system at the acceptable storage and processing overhead. We can deploy a dedicated blockchain on a set of trustworthy nodes rather than operating the blockchain over the entire network. Consortium blockchain is proposed. We also develop efficient sharding-based blockchain protocols to scale the whole blockchain system and make each consensus node only download a part of the transaction data. To efficiently provide massive data to support the Metaverse, on/off-chain data storage mechanisms should be elaborately designed to store the data with high reliability and low redundancy. The lightweight and green consensus protocols tailored to the resource-limited Metaverse systems can be considered. The conventional consensus algorithms such as PBFT can be revised to reduce the mining-related energy consumption. For the flexible interactions among different blockchains, we necessitate a well-designed cross-chain communication scheme to balance the tradeoff between the energy consumption and blockchain throughput.

\subsection{AI Applications for Green Metaverse}
AI provides technology support to build the foundation and promote the development of the Metaverse, but also presents learning based algorithms to improve the decision-making capability for achieving the green Metaverse.  

On one hand, AI technologies are leveraged to build virtual worlds in the Metaverse. Deep learning models are proposed to address the unsupervised learning and supervised learning tasks including scene rendering and creation, object detection and human action recognition. For achieving remarkable energy savings, we further revise those original computation and storage-intensive deep learning models to eliminate the energy consumption. We can adopt the solutions including model compression, parameter pruning and quantization, low-rank approximation and knowledge distillation. By reducing the sizes and complexity of current deep learning models, we can avoid consuming too much energy on the AI training and inference, and find an energy-efficient way to fuse AI with Metaverse. On the other hand, AI algorithms such as deep reinforcement learning are adopted to make intelligent decisions about task partitioning, power control, channel allocation, resource management, user association, movement prediction and data sharing for the optimization problems of the green Metaverse.

\section{Conclusion}
To achieve the green Metaverses, we studied technologies and optimization methods on the MSP side to serve the co-located users in an energy-efficient manner.  We took an AR application as an example, and proposed shared components of the co-located users such that an MSP can utilize the same input/output data and perform the identical workloads once, thereby eliminating redundant data transmission and processing. Since each user was rational to decide whether to accept the offloading service provided by the MSP, we studied how to maximize the economic benefits of the MSP. A Stackelberg game was leveraged for tackling the service reservation and pricing problem. Numerical results indicated that the proposed scheme achieves an advantage in gaining energy savings and satisfying the individual rationality simultaneously. Finally, we identified and discussed open directions of the green Metaverse with the assistance of several enabling technologies.

\section*{Acknowledgments}
This work was supported in part by National Natural Science Foundation of China under Grant 62001125, Grant 62102099, Grant 62003099, and Grant U22A2054,  in part by Guangzhou Basic Research Program under Grant 2023A04J0340, 2023A04J1699, and Grant 2023A04J1704.  Yuan Wu’s work was supported in part by Science and Technology Development Fund of Macau SAR under Grant FDCT 0158/2022/A, and in part by the Guangdong Basic and Applied Basic Research Foundation under Grant 2022A1515011287. Corresponding author: Jianwen Kang and Yuan Wu.

\bibliography{myreference}

\section*{BIOGRAPHIES}
XUMIN HUANG received the Ph.D. degree in control science and engineering from the Guangdong University of Technology, Guangzhou, China, in 2019. He is currently an Associate Professor with the School of Automation, Guangdong University of Technology. He is also currently working as a Macau Young Scholars Postdoctoral Fellow in the State Key Laboratory of Internet of Things for Smart City, University of Macau, Macau, China. His research interests include resource and service optimizations for connected vehicles, Internet of Things, blockchain and edge intelligence.

YUAN WU [S'08, M'10, SM'16] received the Ph.D. degree in Electronic and Computer Engineering from the Hong Kong University of Science and Technology in 2010. He is currently an Associate Professor with the State Key Laboratory of Internet of Things for Smart City, University of Macau, and with the Department of Computer and Information Science, University of Macau. His research interests include green communications and computing, mobile edge computing, and edge intelligence.

JIAWEN KANG  received the M.S. degree and the Ph.D. degree from the Guangdong University of Technology, China, in 2015 and 2018. He is currently a full professor at the Guangdong
University of Technology. He was a postdoc at Nanyang Technological University from 2018 to 2021, Singapore. His research interests mainly focus on blockchain, security, and privacy protection in wireless communications and networking.

JIANGTIAN NIE received the B.Eng. degree in electronics and information engineering from Huazhong University of Science and Technology, Wuhan, China, in 2016, and the Ph.D.
degree from ERI@N, Interdisciplinary Graduate School, Nanyang Technological University (NTU),
Singapore, in 2021. She is currently a Research Fellow with NTU. Her research interests include network economics, game theory, crowdsensing, and learning.

WEIFENG ZHONG received the Ph.D. degree in control science and engineering from the Guangdong University of Technology, Guangzhou, China, in 2019, where he is currently an Associate Professor. He was a visiting scholar with the Nanyang Technological University, Singapore, in 2021, and a visiting student with the Hong Kong University of Science and Technology, Hong Kong, in 2016. His research interests include resource management in smart grid, connected vehicles, and Internet of Things.  
 
DONG IN KIM [M'91, SM'02, F'19]  received the Ph.D. degree in electrical engineering from the University of Southern California, Los Angeles, CA, USA, in 1990. He was a Tenured Professor with the School of Engineering Science, Simon Fraser University, Burnaby, BC, Canada. Since 2007, he has been an SKKU-Fellowship Professor with the College of Information and Communication Engineering, Sungkyunkwan University (SKKU), Suwon, South Korea. He is a Fellow of the Korean Academy of Science and Technology and a Member of the National Academy of Engineering of Korea. He was the Founding Editor-in-Chief for the IEEE Wireless Commun. LETTERS, from 2012 to 2015. He was selected the 2019 recipient of the IEEE Communications Society Joseph LoCicero Award for Exemplary Service to Publications. He is the General Chair for IEEE ICC 2022 in Seoul.
 
SHENGLI XIE [M'01–SM'02–F'19]  received the B.S. degree in mathematics from Jilin University, Changchun, China, in 1983, the M.S. degree in mathematics from Central China Normal University, Wuhan, China, in 1995, and the Ph.D. degree in control theory and applications from South China University of Technology, Guangzhou, China, in 1997. He is currently a Full Professor and the Head of the Institute of Intelligent Information Processing, Guangdong University of Technology, Guangzhou. He has coauthored two books and more than 150 research papers in refereed journals and conference proceedings and was awarded Highly Cited Researcher in 2020. His research interests include blind signal processing, machine learning, and Internet of Things. Dr. Xie received the Second Prize of National Natural Science Award of China in 2009. He is an Associate Editor for IEEE TRANSACTIONS ON SYSTEMS, MAN, AND CYBERNETICS: SYSTEMS. He is a Foreign Full Member (Academician) of the Russian Academy of Engineering.

\end{document}